\documentclass{article}

\usepackage{arxiv}

\usepackage[utf8]{inputenc} 
\usepackage[T1]{fontenc}    
\usepackage{hyperref}       
\usepackage{url}            
\usepackage{booktabs}       
\usepackage{amsfonts}       
\usepackage{nicefrac}       
\usepackage{microtype}      
\usepackage{cleveref}       
\usepackage{lipsum}         
\usepackage{graphicx}
\usepackage{natbib}
\usepackage{doi}
\usepackage{multirow}

\title{Liver Fibrosis Quantification and Analysis: The LiQA Dataset and Baseline Method}

\date{}

\newif\ifuniqueAffiliation
\uniqueAffiliationtrue

\ifuniqueAffiliation 
\author{
Yuanye Liu$^{1}$,
Hanxiao Zhang$^{2}$,
Jiyao Liu$^{3}$,
Nannan Shi$^{4}$,
Yuxin Shi$^{4}$,
Arif Mahmood$^{5}$,
Murtaza Taj$^{6}$,
Xiahai Zhuang$^{1}$ \\
\\
$^{1}$School of Data Science, Fudan University, Shanghai, China \\
$^{2}$Institute of Medical Robotics, Shanghai Jiao Tong University, Shanghai, China \\
$^{3}$Institute of Science and Technology for Brain-Inspired Intelligence, Fudan University, Shanghai, China \\
$^{4}$Department of Radiology, Shanghai Public Health Clinical Center, Shanghai, China \\
$^{5}$Department of Computer Science, Information Technology University, Lahore, Pakistan \\
$^{6}$Department of Computer Science, Lahore University of Management Sciences, Lahore, Pakistan \\
}
\else
\usepackage{authblk}

\newbox{\orcid}\sbox{\orcid}{\includegraphics[scale=0.06]{orcid.pdf}} 
\author[1]{%
	\href{https://orcid.org/0000-0000-0000-0000}{\usebox{\orcid}\hspace{1mm}David S.~Hippocampus\thanks{\texttt{hippo@cs.cranberry-lemon.edu}}}%
}
\author[1,2]{%
	\href{https://orcid.org/0000-0000-0000-0000}{\usebox{\orcid}\hspace{1mm}Elias D.~Striatum\thanks{\texttt{stariate@ee.mount-sheikh.edu}}}%
}
\affil[1]{Department of Computer Science, Cranberry-Lemon University, Pittsburgh, PA 15213}
\affil[2]{Department of Electrical Engineering, Mount-Sheikh University, Santa Narimana, Levand}
\fi

\renewcommand{\shorttitle}{\textit{arXiv} Template}

\hypersetup{
pdftitle={A template for the arxiv style},
pdfsubject={q-bio.NC, q-bio.QM},
pdfauthor={David S.~Hippocampus, Elias D.~Striatum},
pdfkeywords={First keyword, Second keyword, More},
}

\begin{document}
\maketitle

\begin{abstract}
Liver fibrosis represents a significant global health burden, necessitating accurate staging for effective clinical management. 
This report introduces the LiQA (Liver Fibrosis Quantification and Analysis) dataset, established as part of the CARE 2024 challenge. Comprising $440$ patients with multi-phase, multi-center MRI scans, the dataset is curated to benchmark algorithms for Liver Segmentation (LiSeg) and Liver Fibrosis Staging (LiFS) under complex real-world conditions, including domain shifts, missing modalities, and spatial misalignment. 
We further describe the challenge's top-performing methodology, which integrates a semi-supervised learning framework with external data for robust segmentation, and utilizes a multi-view consensus approach with Class Activation Map (CAM)-based regularization for staging. Evaluation of this baseline demonstrates that leveraging multi-source data and anatomical constraints significantly enhances model robustness in clinical settings.
\end{abstract}


\section{Introduction}
The liver serves as the body's largest organ, performing essential synthetic and detoxification functions that are critical for maintaining homeostasis~\citep{J_2019Hepatology_liver_disease}. 
However, chronic liver diseases (CLD) stemming from viral hepatitis, alcohol-related injury, or metabolic dysfunction-associated steatotic liver disease (MASLD), have emerged as a staggering global health burden, affecting approximate $1.7$ billion individuals worldwide~\citep{zhang2025global}.
As CLD progresses, the accumulation of extracellular matrix leads to fibrosis, which serves as a key precursor to cirrhosis and hepatocellular carcinoma (HCC)~\citep{J_2005JCI_liver_fibrosis}.
Consequently, the early and accurate assessment of liver structure and fibrosis severity is paramount for clinical decision-making, enabling timely interventions that can halt or reverse disease progression.

In current clinical workflows, liver biopsy is often regarded as the gold standard for staging fibrosis. Yet, its invasive nature, potential for sampling errors, and associated morbidity limit its utility for longitudinal monitoring~\citep{J_2009Hepatology_liver_biopsy_issue}.
As a result, Magnetic Resonance Imaging (MRI) has become an indispensable non-invasive alternative. MRI offers superior soft-tissue contrast and multi-parametric capabilities, allowing for the comprehensive characterization of liver tissue through diverse sequences such as T2-weighted imaging, diffusion-weighted imaging (DWI), and dynamic contrast-enhanced phases~\citep{J_1997_multiseq_mri}.
Despite these advantages, the quantitative analysis of such complex high-dimensional data remains a bottleneck. Robust computational tools are urgently needed to automate two primary tasks: Liver Segmentation (LiSeg), which is a prerequisite for volumetry and radiomics, and Liver Fibrosis Staging (LiFS), which is essential for objective diagnosis.

To benchmark algorithmic solutions against the rigors of clinical practice, the CARE 2024 challenge (Comprehensive Analysis \& computing of REal-world medical images) was organized in conjunction with MICCAI 2024. Within this initiative, the Liver Fibrosis Quantification and Analysis (LiQA) track was established to address the "real-world" imperfections often absent in curated academic datasets. The Liver Segmentation (LiSeg) task specifically focuses on segmenting the liver from hepatobiliary (HBP) MRI scans. This task presents a dual challenge: first, the HBP phase captures specific functional information that may differ anatomically from structural phases. Second, the task operates in a semi-supervised setting with limited ground truth annotations, exacerbated by significant distribution shifts caused by multi-center and multi-vendor data acquisition.

Beyond anatomical delineation, the LiFS task aims to accurately predict fibrosis stages using multi-sequence MRI scans. While fibrosis is clinically classified into four stages (S1–S4) , the challenge emphasizes two clinically critical binary classifications: staging cirrhosis (S1–3 vs. S4) and identifying substantial fibrosis (S1 vs. S2–4)~\citep{J_2018Rad_LiFS_CT_1}.
Developing automated systems for this task is complicated by the inherent heterogeneity of real-world data. Unlike standardized datasets, the LiQA cohort features random missing modalities, where specific sequences may be absent for certain patients, and spatial misalignments among multi-phase images due to respiratory motion or acquisition variances. These factors require algorithms that are not only accurate but also robust to incomplete and misaligned inputs.

This report presents the LiQA dataset and details the baseline methodology established by the challenge's top-performing team. We describe a comprehensive framework that leverages semi-supervised learning to overcome annotation scarcity and employs a multi-view consensus strategy to integrate heterogeneous MRI sequences. By addressing these complexities, this work aims to establish a solid foundation for reproducible research and advance the deployment of AI in clinical liver analysis.
\section{Dataset}
\label{sec:dataset}

\begin{figure}[t]
    \centering
    \includegraphics[width=\textwidth]{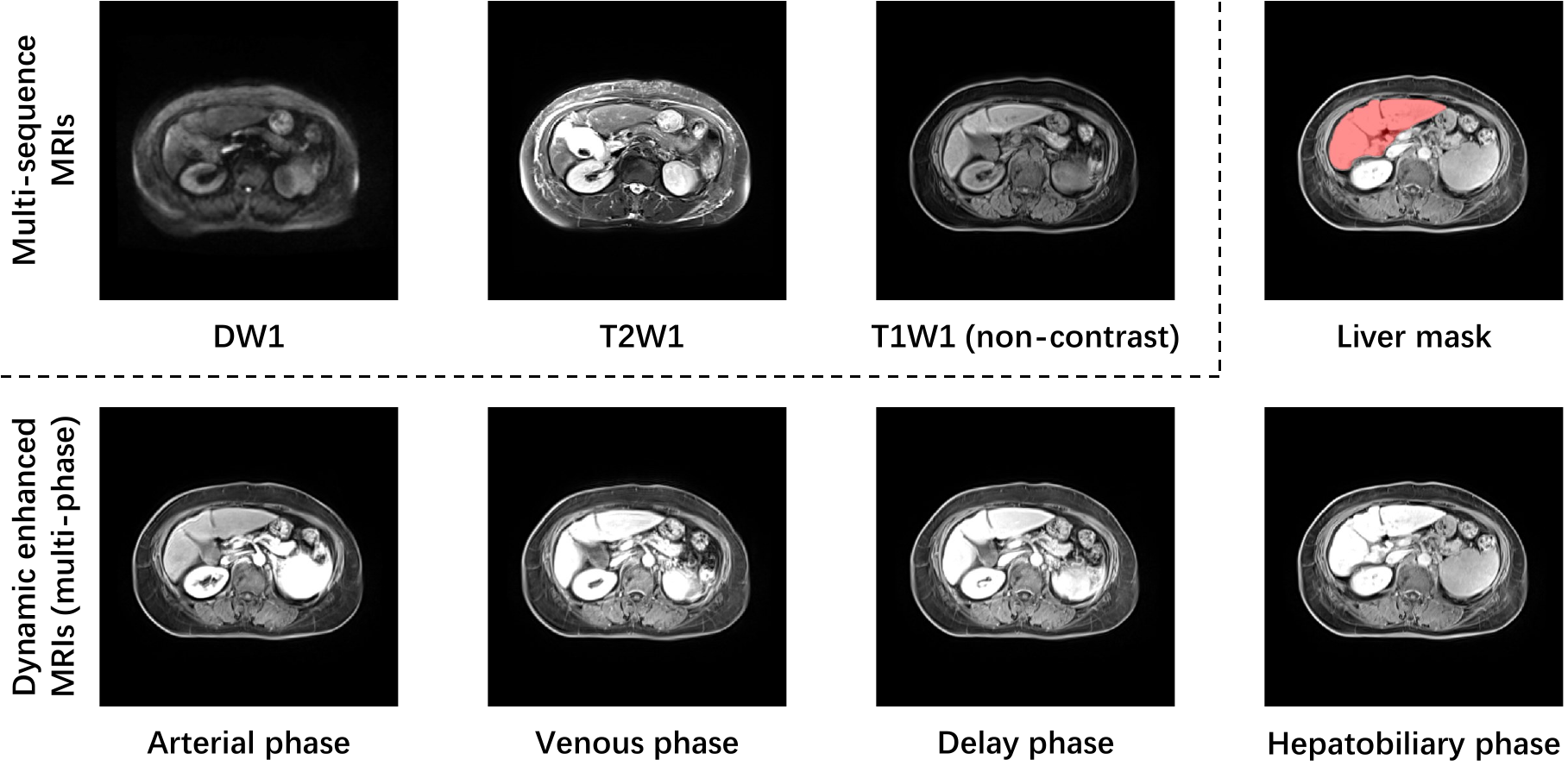}
    \caption{Overview of the LiQA dataset, illustrating the multi-center and multi-phase MRI.}
    \label{fig:intro}
\end{figure}

\subsection{Overview}

We introduce the Liver Fibrosis Quantification and Analysis (LiQA) dataset, a large-scale, open-access benchmark explicitly curated to reflect the complexities inherent in real-world clinical liver imaging. 
The dataset comprises a cohort of $440$ patients diagnosed with liver fibrosis, aggregated from three distinct clinical centers and scanned using diverse multi-sequence, contrast-enhanced MRI protocols. 
In total, the LiQA dataset contains approximately $2,700$ 3D MRI volumes, averaging six or seven sequences per patient, along with nearly $19,000$ slices, offering a substantial resource for training deep learning models. To facilitate broad research participation, the training and validation subsets, complete with expert annotations, are publicly available\footnote{\url{https://zmic.org.cn/care2024/track_3}}.

As detailed in Table~\ref{tab:vendor_partition}, the data collection spans three institutions and utilizes four different MRI scanners from three different vendors, covering both $1.5$T and $3.0$T magnetic field strengths.
This multi-vendor setup is critical for evaluating algorithmic robustness. Vendors A and B (comprising two clinics and three scanners, specifically Siemens Skyra and Philips Ingenia) contributed the majority of the data, with $280$ subjects allocated for training and validation.
To rigorously assess model generalization to unseen domains, Vendor C (Siemens Aera $1.5$T) served exclusively as an out-of-distribution (OOD) test set, providing $40$ patients from an independent scanner that was completely unseen during the training phase. It is important to note that all cases were derived from routine clinical practice; consequently, the dataset captures natural physiological variability and may present co-existing liver lesions, such as hepatocellular carcinoma (HCC).

\begin{table}[t]
\centering
\caption{Data partition statistics across vendors and centers.}
\label{tab:vendor_partition}
{
\begin{tabular}{lllc}
\toprule
\textbf{Vendor} & \textbf{Center \& Scanner} & \textbf{Partition} & \textbf{\#Patients} \\
\midrule
Vendor A  & SPHCC, Siemens Skyra 3.0T & Train / Val / Test (ID) & 100 / 10 / 40 \\
Vendor B1 & SPHCC, Philips Ingenia 3.0T & Train / Val / Test (ID) & 100 / 10 / 40 \\
Vendor B2 & Nantong Hospital, Philips Ingenia 3.0T & Train / Val / Test (ID) & 50 / 10 / 40 \\
Vendor C  & Zhongshan Hospital, Siemens Aera 1.5T & Test (OOD) & 40 \\
\bottomrule
\end{tabular}
}
\end{table}

The MRI acquisition adhered to a standard multi-phase dynamic contrast-enhanced protocol using the Gd-EOB-DTPA agent, which is widely used for liver characterization~\citep{J_2010Radiology_GED_MRI}. 
The imaging dataset includes a comprehensive set of sequences: T2-weighted imaging, diffusion-weighted imaging (DWI), and Gadolinium-enhanced dynamic MRIs. 
Specifically, the dynamic phases capture the temporal evolution of contrast enhancement. Gd-EOB-DTPA (Primovist/Eovist) is administered at a standard dose of 0.1 mL/kg body weight (0.025 mmol/kg) via intravenous bolus injection, typically followed by a saline flush. The arterial phase is acquired approximately 25 seconds post-injection, followed by the portal venous phase at 1 minute, the delayed phase after an additional 90 seconds, and finally the hepatobiliary phase (HBP) acquired 20 minutes post-injection.

Establishing a reliable ground truth is essential for supervised learning. The reference standard for fibrosis staging in this dataset was determined via histopathological examination. Ground-truth labels were derived from biopsy or surgical resection specimens obtained within three months of the MRI scan. This rigorous inclusion criterion ensures high clinical consistency and reliability between the imaging features and the pathological diagnosis.

\subsection{Real-world Design Features}
The LiQA dataset is explicitly designed to replicate the challenges encountered in clinical image analysis, providing a rigorous platform for developing robust algorithms. Key characteristics include:
\begin{itemize}
    \item \textbf{Spatial Misalignment:} The multi-sequence MRI volumes (T1, T2, DWI, and dynamic phases) are not spatially registered, reflecting the routine acquisition variability found in clinical workflows.
    \item \textbf{Missing Modalities:} To simulate real-world data imperfections, certain subjects randomly lack specific sequences due to acquisition failures, motion artifacts, or protocol variations. This feature makes LiQA suitable for evaluating modality-robust learning and imputation methods.
    \item \textbf{Cross-center Heterogeneity:} The inclusion of data from three hospitals and multiple vendors (Siemens, Philips) introduces significant domain shifts. This diversity is critical for testing the cross-site generalization capabilities of deep learning models.
\end{itemize}

\subsection{CARE 2024 Tasks}
The challenge encompasses two primary tasks tailored to liver disease analysis:
\begin{enumerate}
    \item \textbf{LiSeg (Liver Segmentation):} The objective is to automatically segment the liver from single-phase (HBP) MRI scans. This task is designed to address the challenge of "limited annotations", a common bottleneck in medical AI. Ground truth masks are provided for only a small subset (10 cases per center) of the training data, requiring participants to leverage semi-supervised or few-shot learning techniques.
    \item \textbf{LiFS (Liver Fibrosis Staging):} The objective is to accurately predict the fibrosis stage (S1–S4) using multi-phase, multi-center MRI scans. Performance is evaluated on two clinically significant binary classification sub-tasks: staging cirrhosis (S1–3 vs. S4) and identifying substantial fibrosis (S1 vs. S2–4). This task tests the ability of models to integrate complementary information from multiple sequences to form a holistic diagnostic prediction.
\end{enumerate}

\section{Method}
\label{sec:method}
To establish a strong baseline for the LiQA challenge, we reference the winning solution from the challenge participants~\citep{zhang2024care}.
This method addresses the core difficulties of the dataset—specifically the scarcity of annotations for segmentation and the multi-view nature of fibrosis staging—through a semi-supervised segmentation framework and an attention-guided staging pipeline.

\subsection{Liver Segmentation (LiSeg)}
The primary challenge in the LiSeg task is the limited availability of ground truth masks for the hepatobiliary phase (HBP) MRI, coupled with domain shifts across centers. To overcome this, the baseline method employs a semi-supervised learning strategy that leverages both external public datasets and internal unlabeled data.

\subsubsection{Data Augmentation with External Sources}
To augment the training distribution and learn robust anatomical priors, three external open-source datasets are incorporated: CHAOS 2019~\citep{kavur2021chaos}, AMOS 2022~\citep{ji2022amos}, and ATLAS 2023~\citep{quinton2023tumour}. These datasets provide liver masks across multiple MRI sequences (T1, T2, DWI, and CE-MRI), compensating for the single-phase nature of the labeled LiQA data.

\begin{figure}[t]
\centering
\includegraphics[width = \textwidth]{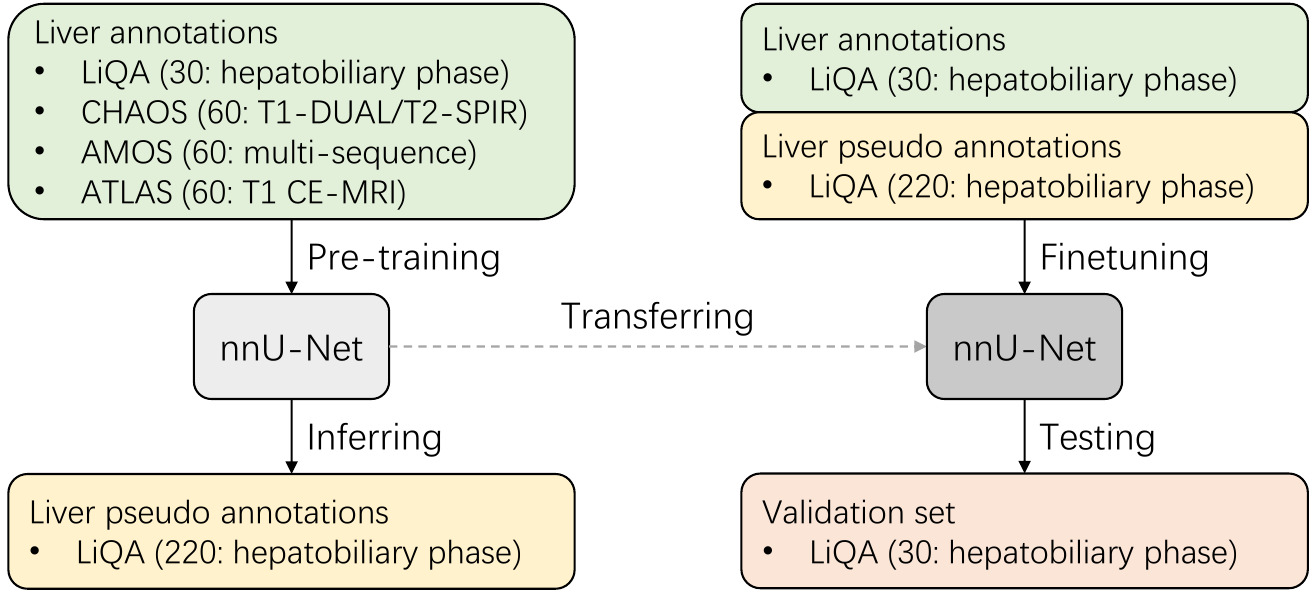}
\caption{Framework for liver segmentation method that utilizes both labeled and unlabeled data.
}
\label{fig:framework} 
\end{figure}

\subsubsection{Semi-supervised Pseudo-labeling Framework}
The segmentation pipeline utilizes the nnU-Net architecture~\citep{J_2021NatureMethods_nnUNet} and follows a three-stage training process as follows, shown in Fig.~\ref{fig:framework},
\begin{enumerate}
    \item \textbf{Pre-training:} A model is first pre-trained from scratch for $1,000$ epochs using the combined external labeled datasets to establish a generalizable feature representation.
    \item \textbf{Pseudo-label Generation:} The pre-trained model is applied to the unlabeled HBP images in the LiQA training set, generating pseudo-masks for $220$ previously unlabeled cases.
    \item \textbf{Fine-tuning:} The final model is fine-tuned for $300$ epochs on a composite dataset containing the manually labeled LiQA data ($30$ cases), the pseudo-labeled LiQA data ($220$ cases), and the external data.
\end{enumerate}
This strategy effectively expands the supervision signal while adapting the model to the specific characteristics of the LiQA HBP sequence.

\subsection{Liver Fibrosis Staging (LiFS)}
\begin{figure}[t]
\centering
\includegraphics[width=\textwidth]{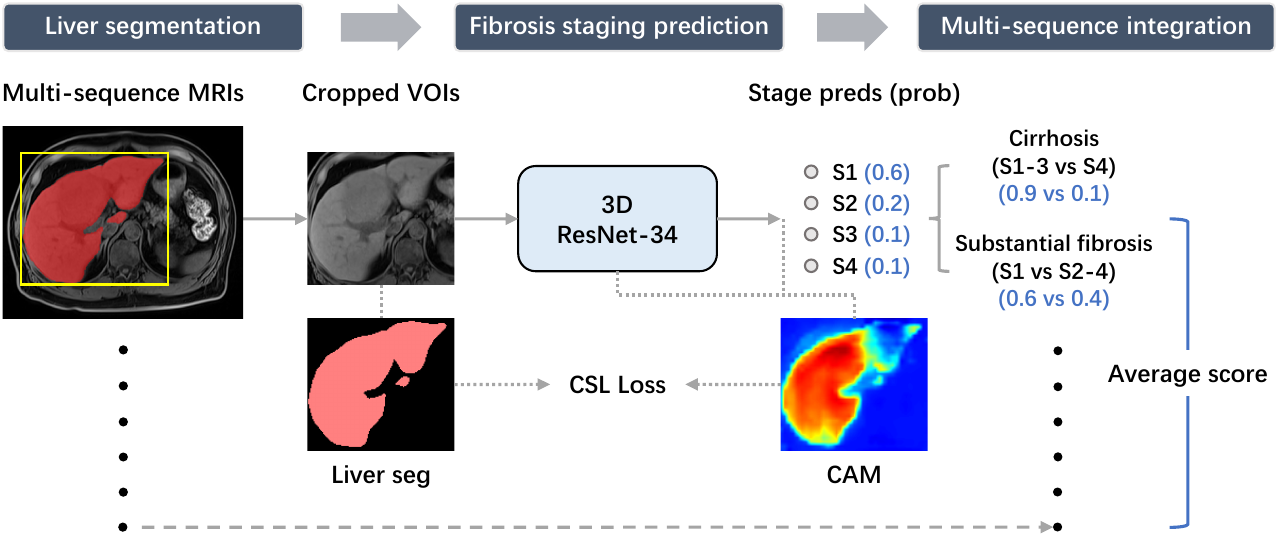}
\caption{Framework for liver fibrosis staging pipeline based on multi-sequence MRIs.
}
\label{fig_framework_LiFS} 
\end{figure}

The LiFS task requires integrating information from multi-sequence MRIs that may be spatially misaligned or randomly missing. The baseline method proposes a three-step pipeline: Volume of Interest (VOI) extraction, fibrosis scoring, and multi-view fusion. The framework is shown in Fig.~\ref{fig_framework_LiFS}.

\subsubsection{VOI Extraction and Regularization}
First, the pre-trained LiSeg model is used to crop liver VOIs from each available MRI sequence for a patient. 
These VOIs serve as input to a 3D ResNet-34 backbone~\citep{he2016deep}.
To ensure the classification model focuses on relevant liver tissue rather than background artifacts, the method employs CAM-SEM Loss (CSL) regularization~\citep{zhang2023trustworthy}.
This technique constrains the online-generated Class Activation Maps (CAMs) to align with the liver segmentation masks, forcing the network to attend specifically to the liver region during training~\citep{zhou2016learning}.

\subsubsection{Multi-view Fusion}
Since different patients have different combinations of available sequences, a flexible fusion strategy is required. 
The model predicts a fibrosis score for each individual MRI sequence. 
The final patient-level staging is derived by averaging the predicted scores from all available sequences.
This consensus approach mitigates the impact of missing modalities and noise in individual sequences. 
The baseline explores multiple formulations for the scoring task, including binary classification, 4-category classification, and ordinal regression.

\section{Results}
\label{sec:result}

\subsection{Liver Segmentation Results}
Performance for the LiSeg task is evaluated using Dice Similarity Coefficient (DSC) and Hausdorff Distance (HD).

\textbf{Validation Performance:}
The impact of the semi-supervised strategy was analyzed on the validation set. 
A baseline model trained solely on the $30$ labeled LiQA cases achieved a DSC of $0.9598$ and an HD of $39.14$ mm. 
Direct fine-tuning with pseudo-labels initially degraded the HD to $51.66$ mm due to noise in the pseudo-masks. 
However, incorporating external data stabilized the process. The final model (LiQA labeled + Pseudo-labeled + External) achieved the best performance with a \textbf{DSC of $0.9621$} and an \textbf{HD of $35.99$ mm}.
This indicates that external multi-sequence data is critical for correcting domain gaps in the HBP-only LiQA dataset.

\textbf{Test Performance:}
On the unseen test set, the submitted method achieved a \textbf{DSC of $0.9350$ $\pm$ $0.0970$} and an \textbf{HD of $36.74$ $\pm$ $18.47$ mm}. These results demonstrate strong generalization capabilities to unseen data distributions.

\subsection{Liver Fibrosis Staging Results}
The LiFS task is evaluated based on the Area Under the Curve (AUC) and Accuracy (ACC) for two clinical sub-tasks: Cirrhosis discrimination (S1-3 vs. S4) and Substantial Fibrosis discrimination (S1 vs. S2-4).

\textbf{Validation Performance:}
A comparison of different staging heads (Cross-Entropy, Binary Cross-Entropy, Ordinal Regression) on the validation set is summarized in Table \ref{tab:lifs_val}. The Ordinal Regression approach yielded the highest AUC scores of \textbf{0.8108} (Cirrhosis) and \textbf{0.8056} (Substantial Fibrosis). The Cross-Entropy method combined with CSL regularization (CE wCSL) provided the most balanced performance across metrics, achieving the highest accuracy for identifying substantial fibrosis (0.8000).

\textbf{Test Performance:}
For the final submission on the test set, the CE wCSL method was selected for its stability. It achieved an AUC of \textbf{$0.6481$ $\pm$ $0.1072$} for Cirrhosis and \textbf{$0.5931$ $\pm$ $0.2084$} for Substantial Fibrosis. While the drop in AUC suggests challenges with out-of-distribution generalization (Vendor D), the accuracy remained relatively stable at approximately $71-73\%$.

\begin{table}[t]
\centering
\caption{Quantitative performance of different methods on the LiQA validation set for LiFS task.}
\label{tab:lifs_val}
\begin{tabular}{lllll}
\hline\hline
\multirow{3}{*}{Method \quad \quad } & \multicolumn{2}{l}{Cirrhosis} & \multicolumn{2}{l}{Substantial fibrosis} \\
 & \multicolumn{2}{l}{(S1-3 vs S4)} & \multicolumn{2}{l}{(S1 vs S2-4)} \\ \cline{2-5} 
 & AUC $\uparrow$ \quad \quad & ACC $\uparrow$ \qquad \qquad & AUC $\uparrow$ \quad \quad & ACC $\uparrow$ \qquad  \\ \hline
CE & 0.7498 & 0.6000 & 0.6111 & 0.7333 \\
BCE & 0.7790 & 0.7000 & 0.6667 & 0.7000 \\
BCE wCSL & 0.8083 & 0.6333 & 0.6852 & 0.7333 \\
Reg & 0.7181 & 0.7000 & 0.7639 & 0.7000 \\
Ord \citep{diaz2019soft} & \textbf{0.8108} & 0.6000 & \textbf{0.8056} & 0.7667 \\
CE wCSL & 0.7683 & \textbf{0.7000} & 0.7523 & \textbf{0.8000} \\ \hline\hline
\end{tabular}%
\end{table}

\section{Conclusion}
\label{sec:conclusion}
The CARE 2024 LiQA track successfully established a benchmark for liver fibrosis analysis using real-world clinical data. The dataset's inclusion of multi-center, multi-vendor, and misaligned scans provides a rigorous testbed for algorithmic generalization. The winning method demonstrated that semi-supervised learning with external data is essential for segmentation when annotations are scarce. Furthermore, multi-view consensus combined with attention-guided regularization (CSL) offers a promising path for robust disease staging in the presence of missing or heterogeneous data. Future work should focus on developing interpretable fusion rules and integrating multi-modal data (e.g., CT) to further enhance clinical applicability.

\bibliographystyle{unsrtnat}
\bibliography{strings,references}







\end{document}